\pdfoutput=1
\documentclass[11pt]{article}

\usepackage{EMNLP2023}

\usepackage{times}
\usepackage{latexsym}

\usepackage{todonotes}
\usepackage{amsmath}

\usepackage{optidef}
\usepackage{amsfonts}

\usepackage[T1]{fontenc}
\usepackage[utf8]{inputenc}

\usepackage{microtype}

\usepackage{inconsolata}
\usepackage{booktabs}
\usepackage{multirow}
\usepackage{array}
\usepackage{varwidth}
\newcolumntype{M}{>{\begin{varwidth}{16cm}}l<{\end{varwidth}}} 
\newcolumntype{E}{>{\begin{varwidth}{8cm}}l<{\end{varwidth}}} 
\newcolumntype{N}{>{\begin{varwidth}{5cm}}l<{\end{varwidth}}} 
\newcolumntype{P}[1]{>{\centering\arraybackslash}p{#1}}
\usepackage{soul,color}

\newcommand{\llm}[1]{#1LLMs}
\newcommand{\hide}[1]{}

\usepackage{amstext} % for \text macro
\usepackage{array}   % for \newcolumntype macro
\newcolumntype{L}{>{$}l<{$}} % math-mode version of "l" column type
\newcolumntype{C}{>{$}l<{$}}

\definecolor{lpink}{cmyk}{0, 0.7808, 0.4429, 0.1412}
\definecolor{aqua}{cmyk}{0.91, 0, 0.09, 0.36}
\definecolor{ao}{rgb}{0.0, 0.5, 0.0}
\definecolor{amber}{rgb}{1.0, 0.49, 0.0}
\definecolor{dblue}{rgb}{0.0, 0.0, 0.61}
\definecolor{burgundy}{rgb}{0.5, 0.0, 0.13}
\definecolor{dgreen}{rgb}{0.0, 0.61, 0.0}
\definecolor{purple}{rgb}{0.61, 0.0, 0.5}
\definecolor{randompink}{rgb}{0.859,0.56,0.884}
\definecolor{entropyblue}{rgb}{0.0,0.288,0.678}

\definecolor{kmy-color}{rgb}{0.858, 0.188, 0.478}

\definecolor{blueshade}{HTML}{81affe}

\definecolor{orangeshade}{HTML}{ffd8a7}

\newcommand{\usc}{$^\ddag$}
\newcommand{\nus}{$^\dagger$}
\newcommand{\astar}{$^\S$}
\newcommand{\stanford}{$^\P$}

\title{CoAnnotating: Uncertainty-Guided Work Allocation between \\Human and Large Language Models for Data Annotation}
\author{Minzhi Li \nus \astar \hspace{1.5em}
        Taiwei Shi \usc \hspace{1.5em}
        Caleb Ziems \stanford \hspace{1.5em}\\
        \bf Min-Yen Kan \nus \hspace{1.5em}
        \bf Nancy F. Chen \astar \hspace{1.5em}
        \bf Zhengyuan Liu \astar \hspace{1.5em}
        \bf Diyi Yang \stanford \hspace{1.5em}\\
        \nus National University of Singapore \hspace{1.5em}
        \astar Institute for Infocomm Research (I$^2$R), A*STAR \\
        \usc University of Southern California \hspace{1.5em}
        \stanford Stanford University\\
        \texttt{\href{mailto://li.minzhi@u.nus.edu}{li.minzhi@u.nus.edu}} \hspace{1.5em}
        \texttt{\href{mailto://taiweish@usc.edu}{taiweish@usc.edu}}  \hspace{1.5em} 
        \texttt{\href{mailto://cziems@stanford.edu}{cziems@stanford.edu}}\\
        \texttt{\href{mailto://nfychen@i2r.a-star.edu.sg}{nfychen@i2r.a-star.edu.sg}}  \hspace{1.5em}
        \texttt{\href{mailto://liu\_zhengyuan@i2r.a-star.edu.sg}{liu\_zhengyuan@i2r.a-star.edu.sg}}  \hspace{1.5em}\\
        \texttt{\href{mailto://kanmy@comp.nus.edu.sg}{kanmy@comp.nus.edu.sg}}  \hspace{1.5em}
        \texttt{\href{mailto://diyiy@cs.stanford.edu}{diyiy@cs.stanford.edu}} 
}

\begin{document}
\maketitle
\begin{abstract}
Annotated data plays a critical role in Natural Language Processing (NLP) in training models and evaluating their performance. Given recent developments in Large Language Models (LLMs), models such as ChatGPT demonstrate zero-shot capability on many text-annotation tasks, comparable with or even exceeding human annotators. Such LLMs can serve as alternatives for manual annotation, due to lower costs and higher scalability. However, limited work has leveraged LLMs as complementary annotators, nor explored how annotation work is best allocated among humans and LLMs to achieve both quality and cost objectives. We propose \textit{CoAnnotating}, a novel paradigm for Human-LLM co-annotation of unstructured texts at scale. Under this framework, we utilize uncertainty to estimate \llm{}' annotation capability. 
Our empirical study shows \textit{CoAnnotating} to be an effective means to allocate work from results on different datasets, with up to 21\% performance improvement over random baseline. For code implementation, see \url{https://github.com/SALT-NLP/CoAnnotating}. 
\end{abstract}

\section{Introduction}
Labeled data plays a critical role in establishing benchmarks and developing models for Natural Language Processing (NLP). Although Large Language Models (LLMs) like ChatGPT have demonstrated their strong zero-shot performance in various tasks such as question answering, reasoning, natural language inference,
 sentiment analysis, and named entity recognition, results obtained by fine-tuned language models still outperform \llm{} on most of these tasks \citep{qin2023chatgpt, zhong2023can, ziems2023can}. Therefore, collecting labeled data for model training and fine-tuning is still valuable.
Instead of deploying \llm{} directly for downstream uses, it is worthwhile to investigate how researchers can leverage \llm{}' zero-shot capability in labeling text data to construct high-quality datasets and improve the performance of fine-tuned models.

\begin{figure}[t!]
    \centering \includegraphics[width=\linewidth]{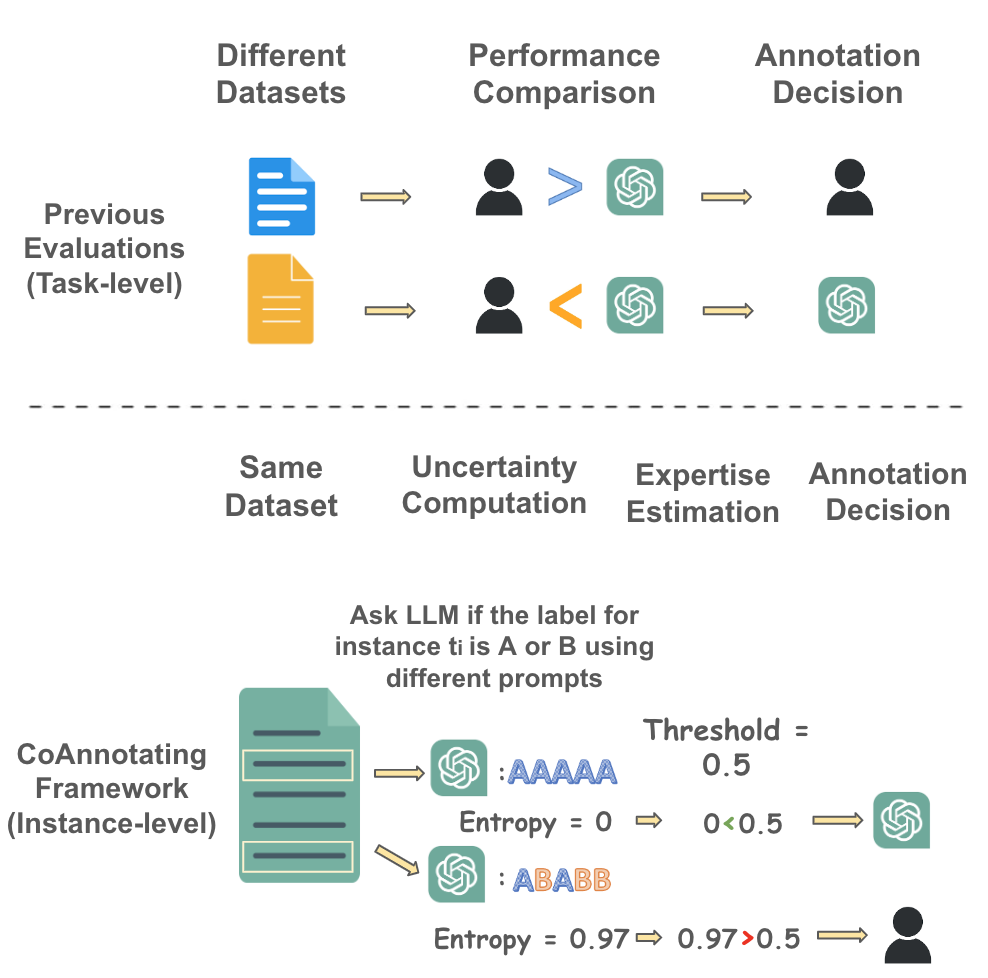}
    \caption{\small \textit{CoAnnotating} framework. It differs from previous work by considering how to allocate data within the \textbf{same} dataset to humans and ChatGPT by obtaining responses from ChatGPT using different variations of prompts and estimating ChatGPT's annotation expertise with the use of uncertainty metrics such as entropy. }
    \label{fig:crown_jewel}
\end{figure}

Typically, researchers recruit human annotators such as experts or crowd workers to perform data annotation \citep{kittur2008crowdsourcing, snow2008cheap}. Some challenges in manual annotation includes high costs of recruiting and training annotators, annotation inconsistency and human subjectivity \citep{lingren2014evaluating, grosman2020eras}. Recent work explored how \llm{} perform relative to crowd workers \citep{ding2022gpt} and results showed that it is possible for \llm{} like ChatGPT to replace large-scale manual annotation \citep{huang2023chatgpt, kuzman2023chatgpt}. In some cases, \llm{}' annotation quality even outperforms human annotators on certain tasks \citep{gilardi2023chatgpt}. Given the much lower annotation cost than crowd workers, \llm{} are considered to have great potential to increase the cost efficiency of the data annotation process. However, some studies also show that, relative to human performance, \llm{}' zero-shot performance falls short on more difficult and pragmatic tasks \citep{wang2021want,kocon2023chatgpt}. They suggest that practitioners should use caution when using \llm{} to annotate data \citep{reiss2023testing, huang2023chatgpt}. Such prior works view humans and \llm{} as \textbf{competitors}, measuring the accuracy of LLM labels as a replacement for human annotation, rather than considering how humans and \llm{} might \textbf{collaborate} in an efficient manner. It is Human-LLM collaboration that motivates this work. We propose the \textit{CoAnnotating} framework, which aims to balance the complementary profiles of humans and \llm{} in terms of their respective annotation quality and cost. 
Our work tackles the problem of Human-LLM co-annotation from a \emph{resource allocation} perspective. Following \citet{gentile2022fast}, \citet{diao2023active} and \citet{wang2021want}, we consider model confidence as a reliable signal for the model's expected performance. As we consider allocating a given datapoint for an LLM to annotate, we can use the inverse of the model's uncertainty to estimate our confidence in that allocation. Under \textit{CoAnnotating}, we quantify \llm{}' annotating expertise on the \textbf{instance-level} (estimating how well \llm{} can annotate the given data point) beyond \textbf{task-level} (evaluating how \llm{} performs on overall for each dataset). As such, a more informed allocation decision can be made with this fine-grained and contextualized instance-level perspective, rather than broad and coarse dataset-level expertise. 

We show that our proposed method using the uncertainty of responses can achieve a more efficient and more accurate work allocation than the random allocation baseline. Our results also show that confidence scores generated by \llm{} are generally well-calibrated but not always reliable. It is possible to outsource some annotation work to achieve human-level performance for more straightforward tasks like topic understanding. On the other hand, a tradeoff between annotation quality and annotation cost is inevitable for more nuanced tasks. Our framework establishes a guide to effectively allocate AI and human efforts in collaborative annotation, and in doing so, it provides key insights into the capacities of \llm{}, as well as the nature of the tasks and data that remain outside these capacities.

\section{Related Work}
\subsection{Weak Supervision}
In a traditional supervised learning setting, every training data point is labeled by human annotators. However, acquiring manually annotated labels for training data can be prohibitively costly and time-consuming. Weak supervision helps to address the challenge using partially and imperfectly labeled data for training \citep{zhang2022survey}. Weak supervision techniques obtain these noisy labels by tapping into heuristics \citep{ratner2017snorkel, meng2018weakly, awasthi2020learning}, feature annotation \citep{mann2010generalized}, external knowledge bases \citep{hoffmann2011knowledge, min2013distant}, pretrained models \citep{bach2019snorkel, zhang2021creating} and third-party tools \citep{lison2020named}. Moreover, weak supervision can be combined with the active learning framework \citep{gonsior2020weakal} to select the most informative data to be annotated by humans and utilize weak supervision to decide noisy labels. Given \llm{}' stunning zero-shot capabilities, our work explores the possibility of using them as a more efficient labeling source, thus freeing up resources to be reinvested in the research pipeline.

\subsection{LLMs for Annotation}
Most prior works frame the decision for human or LLM annotation as one of competition rather than collaboration between these modes. These show that \llm{} like GPT-3 \texttt{davinci-003}  have strong zero-shot sentiment analysis performance \citep{ding2022gpt}. ChatGPT (\texttt{gpt-3.5-turbo}) performs surprisingly well on automatic genre detection in under-resourced languages like Slovenian \citep{kuzman2023chatgpt}. ChatGPT can even achieve high accuracy on some of the most nuanced tasks like implicit hate speech detection \citep{huang2023chatgpt}. Similarly, GPT-4 is able to annotate texts that require reasoning and contextual knowledge and provide explanations that could facilitate interpretive research \citep{tornberg2023chatgpt}. These results show the great potential of LLMs as data annotation tools with just simple prompt design and without much manual labeling efforts
\citep{kuzman2023chatgpt}.

However, there is still room to close significant performance gaps between LLMs' performance and existing fine-tuned baselines on some challenging tasks. 
\llm{} struggle with named entity recognition \citep{ding2022gpt, qin2023chatgpt}, relational reasoning \citep{bang2023multitask}, affective tasks \citep{kocon2023chatgpt, amin2023will} and semantic similarity tasks \citep{kocmi2023large, wang2023chatgpt}. Moreover, it does not outperform fine-tuned baselines for generation tasks like question answering and text summarization \citep{tan2023evaluation, wang2023chatgpt}. These works all take the perspective that LLMs and humans are competitors, making task-level comparisons between LLMs and humans/fine-tuned models for each dataset. Our work views LLMs and humans as potential collaborators, with the possibility to work with each other to annotate the same dataset.

\subsection{Human-Machine Collaboration for Dataset Creation}
The quality of the dataset and the cost of creating a dataset are two important but sometimes conflicting objectives in dataset creation. Previous work suggests a human-AI collaborative framework that utilizes language models' generation capability and human revision and evaluation skills \citep{tekiroglu2020generating, yuan2021synthbio, bartolo2021models, liu2022wanli} to create valuable datasets of high quality. 
For cost efficiency, some have proposed averaging or majority vote over human and machine outputs \citep{chaganty2018price, ziems2023can} and some initial empirical explorations such as analyzing the random combination of distillation of LLM and manual annotation~\citep{kang2023distill} as well as active labeling assignments via the logit outputs \citep{wang2021want}. Our framework takes both quality and cost into consideration by using uncertainty metrics to make informed human-AI work-allocation decisions to ensure cost efficiency without compromising quality.

\section{CoAnnotating Framework}
\begin{figure*}
  \includegraphics[width=\textwidth]{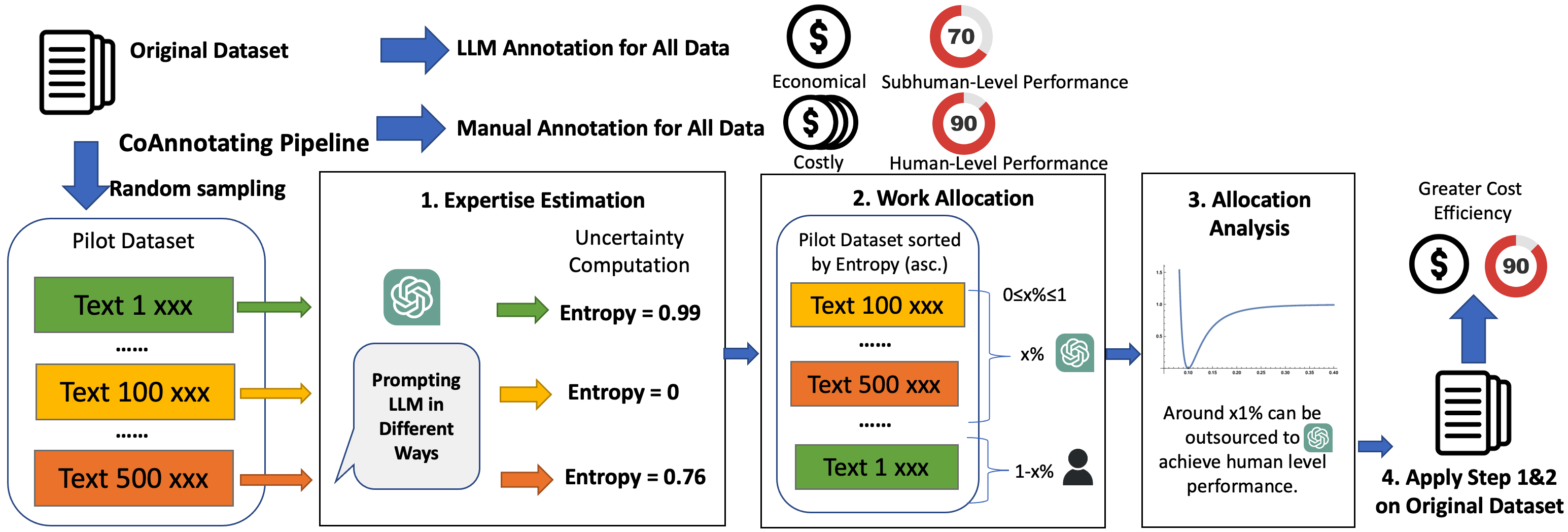}
  \caption{Workflow of \textit{CoAnnotating}. The framework consists of uncertainty-guided expertise estimation, work allocation, and cost performance Pareto analysis. With insights gained from Pareto analysis on the pilot dataset, uncertainty-guided work allocation can be applied on the original unlabeled dataset to achieve greater cost efficiency.}
  \label{fig:pipeline}
\end{figure*}

Our \textit{CoAnnotating} framework sets up a guide for annotating text data collaboratively (Figure \ref{fig:pipeline}). For a given unlabeled train dataset $D_{t}=\{t_1,t_2,...t_m\}$ where $t_i$ is the i-th instance in the dataset, our framework automatically decides whether each data instance should be annotated by human or by \llm{the } (Section \ref{allocation}) by computing the uncertainty level of \llm{the }'s annotations for each instance (Section \ref{estimation}), with the goal of achieving a higher annotation quality and a lower annotation cost for a given dataset (Section \ref{strategy}). 
\subsection{Prompt Construction}
Previous work shows that \llm{}' performance can be highly sensitive to perturbations in input \citep{jang2023consistency}. Therefore, we introduce a set of diverse types of prompts $P_i=\{p_{i1},p_{i2},...,p_{ik}\}$ for each instance $t_i$. Besides the (1) basic instruction format, we vary the prompts by swapping its sequence of sentences (2; \textsl{symmetric perturbation}), paraphrasing the instruction (3; \textsl{semantic perturbation}), enquiring in various question formats (4; \textsl{True/False}, 5; \textsl{Textual Short Responses} 6; \textsl{Multiple Choice Question}) and asking with confirmation bias (7; \textsl{negation perturbation}). \hide{With $k$ distinct prompts $P_i=\{p_{i1},p_{i2},...,p_{ik}\}$ that are carefully crafted for each instance $t_i$, it ensures the robustness of our proposed framework.}
\begin{table}[t]
\centering
\resizebox{7.5cm}{!}{
\fontsize{10.08pt}{10.08pt}\selectfont
\begin{tabular}{@{}p{5.5cm}P{2.5cm}@{}}
\toprule
\multicolumn{2}{l}{\begin{tabular}[c]{@{}l@{}}Text =  Sentence1: \{sentence1\}\\             Sentence2: \{sentence2\}\end{tabular}}                                                                                          \\ \midrule
\multicolumn{1}{c}{\textbf{Prompt}}                                                                                                                                                        & \textbf{Type}                   \\ \midrule
\begin{tabular}[c]{@{}p{5.5cm}@{}}Please label if the following two sentences are paraphrases of each other. Please give your answer as ``paraphrase'' or ``not paraphrase''.\\ \{Text\}\end{tabular} & Instruction                     \\ \midrule
\begin{tabular}[c]{@{}p{5.5cm}@{}}\{Text\}\\ Please label if the two sentences above are paraphrases of each other. Please give your answer as ``paraphrase'' or ``not paraphrase''.\end{tabular}     & Sequence Swapping               \\ \midrule
\begin{tabular}[c]{@{}p{5.5cm}@{}}Given the following two sentences, please classify the relationship of the following two sentences as “paraphrase” or “not paraphrase”.\\ \{Text\}\end{tabular} & Paraphrase                      \\ \midrule
\begin{tabular}[c]{@{}p{5.5cm}@{}}Is it true that the following two sentences are/are not paraphrases of each other? Give your answer as “true” or “false”.\\ \{Text\}\end{tabular}               & True/False                      \\ \midrule
\begin{tabular}[c]{@{}p{5.5cm}@{}}What relationship do the following two sentences have? Is it “paraphrase” or “not paraphrase”?\\ \{Text\}\end{tabular}                                          & Question Answering              \\ \midrule
\begin{tabular}[c]{@{}p{5.5cm}@{}}Please choose one option that best describes the relationship between the following two sentences.\\ \{Text\}\\ (A) Paraphrase\\(B) Not paraphrase\end{tabular}   & Multiple Choice Question        \\ \midrule
\begin{tabular}[c]{@{}p{5.5cm}@{}}I think the following two sentences are/are not paraphrases of each other. Do you agree?\\ \{Text\}\end{tabular}&Question with Confirmation Bias \\ \bottomrule
\end{tabular}}
\caption{Examples of our 7 designed prompt types asking ChatGPT to annotate each instance for the concrete task of paraphrase detection.}
\label{tab:prompt}
\end{table}
\subsection{Uncertainty Computation}
\label{estimation}
In a real-world setting, there is no gold data on which to gauge the model's expected accuracy and thus decide on the optimal annotation strategy.=
However, model confidence can serve as a reliable signal for model performance \citep{gentile2022fast, diao2023active, wang2021want}. Therefore we compute the LLM uncertainty $u_i$ to guide the work-allocation process. We compute $u_i$ in two ways which are easy to implement and have proven effectiveness in previous literature \citep{diao2023active}:=
(1) self-evaluation and (2) entropy. In each case, for $t_i$ by prompting \llm{} $k$ times with different prompts in $P_i$ we get $k$ annotations $A_i=\{a_{i1},a_{i2},...,a_{ik}\}$ for each instance. As an ablation study (\ref{ablation}), we also prompt \llm{} k times with the same prompt to get $k$ annotations to study the effect of prompt perturbations.
\paragraph{Self-Evaluation.} 
Previous work shows that LLMs are well calibrated and can provide information about their uncertainty themselves~\citep{wang2021want, kadavath2022language, diao2023active}. We ask the model to directly output its confidence score \citep{lin2022teaching} by postpending the phrase \textit{"and please give a confidence score on a scale of 0 to 1 for your prediction"}. 
The uncertainty for $t_i$ is calculated by:

\begin{equation*}
    \label{eq:objective}
    u_i=1-\frac{1}{k} \sum_{j=1}^k P_\theta(a_{ij}|p_{ij})
\end{equation*}
%\paragraph{Disagreement}(optional)
where $P_\theta(a_{ij}|p_{ij})$ is the probability of a class label being annotated by ChatGPT given the prompt $p_{ij}$. We obtain its value by extracting the confidence score provided by \llm{} directly. 

\paragraph{Entropy.}
Entropy is a measure of the impurity in a set of data and can be used to quantify the uncertainty associated with the class labels. The larger the entropy value, the more uncertain the responses are. 
We can use this metric to estimate the uncertainty level:
\begin{equation*}
    \label{eq:objective}
    u_i=- \sum_{j=1}^k P_\theta(a_{ij}|p_{ij})\ln P_\theta(a_{ij}|p_{ij})
\end{equation*}
where $P_\theta(a_{ij}|p_{ij})$ is calculated as the frequency of a certain prediction among all predictions.

\subsection{Work Allocation Strategies}
\label{allocation}
Building upon the aforementioned uncertainty level estimation, we can then use the uncertainty level $u_i$ to guide the work allocation. 

\paragraph{Random Allocation.} Random allocation is chosen as a baseline strategy for comparison. This is the strategy that randomly samples $n$ instances ($0\leq n \leq m$) in $D_t$ to be annotated by \llm{} while the remaining $m-n$ data is annotated by humans.

\paragraph{Self-Evaluation Guided Allocation.} \citet{wang2021want} introduces an active label assignment approach that ranks outputs by their logits. Not all LLM APIs support this computation, so we modify this baseline with our self-evaluation approach, 
sorting instances by the self-reported confidence scores in decreasing order. We then select the top $n$ instances ($0\leq n \leq m$) in $D_t$ with the lowest level of uncertainty as the best candidates for LLM annotation. The remaining $m-n$ data points are allocated to human annotators.

\paragraph{Entropy Guided Allocation.} 
It is not possible to entirely ensure the reliability of black box \llm{} self-reported confidence. Therefore, we also propose the use of entropy across \llm{}' responses to gauge their certainty and reliability. We sort the instances by their respective entropy values in increasing order and select the top $n$ instances ($0\leq n \leq m$) in $D_t$ with the lowest level of uncertainty to be annotated by \llm{}. 
Again, the remaining $m-n$ data points with inconsistent responses will be allocated for human annotation.

\subsection{Strategy Selection}
\label{strategy}
We frame the co-annotation process as a multi-objective optimization problem with two main objectives, maximizing annotation quality and minimizing annotation cost. We can determine annotation quality by the classification performance of a model fine-tuned using a certain co-annotation strategy. The total annotation cost is the sum of manual annotation costs and those incurred by the LLM. 
Inspired by \citet{kang2023distill}, we apply the Pareto efficiency concept in strategy selection. Here, the Pareto efficient scenario refers to the situation where it is impossible to increase the classification performance of the fine-tuned model without incurring a higher annotation cost. By adopting different allocation strategies and setting different proportions of data allocated to \llm{}, we get various allocation patterns with different annotation qualities and costs. We can then plot the performances of each quality-cost combination and approximate the Pareto frontier by interpolating the discrete data points \cite{Abdolrashidi_2021_CVPR, treviso-etal-2022-predicting}. 
Practitioners can plot annotation quality against the cost for pilot data to gain a better understanding of this tradeoff, and they can use the Pareto efficient points to decide which ratio of data they should outsource to \llm{} at their desired budget level. 
\section{Experiments}
\subsection{Datasets}

We use six classification datasets for different types of tasks. Since LLM inference costs much less than a human salary, we know the simple allocation decision is to choose \llm{} over humans whenever an LLM achieves a utility greater than or equal to that of human annotators. For a more challenging setting, we identify tasks in which \llm{} are known to struggle with discriminating the underlying constructs   \citep{pikuliak_chatgpt_survey, wang2021want}. 
In such cases, there is a tradeoff between annotation quality and annotation cost and  \textit{CoAnnotating} facilitates better decision-making in such contexts.  If the size of the train data is too large, we will take a stratified random sampling for approximately 1000 samples.  

\paragraph{Topic Classification} is a challenging task for large pretrained language models like GPT-3 \citep{wang2021want}. We choose two representative datasets:
TREC \citep{li2002learning} and AG News \citep{zhang2015character}. AG News contains news titles and their descriptions, which were gathered by an academic news search engine, and which span four topics: \textsl{ world, sports, business, and science/technology}. TREC contains of English questions with six manually labeled class labels:  \textsl{abbreviation; entity; description and abstract concept; human being; location;} and \textsl{numeric value}.

\paragraph{Semantic Similarity} is known to challenge ChatGPT \citep{jang2023consistency}. 
We select MRPC \citep{dolan2005automatically} and TempoWiC \citep{loureiro2022tempowic} as two representative datasets for semantic similarity understanding. MRPC is a corpus of sentence pairs extracted from online news and annotated by humans for whether the sentences are semantically equivalent. TempoWiC contains annotated tweet pairs for whether there is a meaning shift of the target word. 
\paragraph{Nuanced Comprehension} 
We also experiment with Tweet Stance Detection \citep{mohammad2016semeval} and Conversation Gone Awry \citep{zhang2018conversations} to explore the collaboration paradigm on tasks requiring more nuanced comprehension. Tweet Stance Detection in SemEval-2016 \citep{mohammad-etal-2016-semeval} is a dataset of tweets annotated with the author's stance (favorable, neutral, and negative) toward a certain topic and we select the topic of abortion. 
\begin{figure*}[t]
  \includegraphics[width=\textwidth,height=7.8cm]{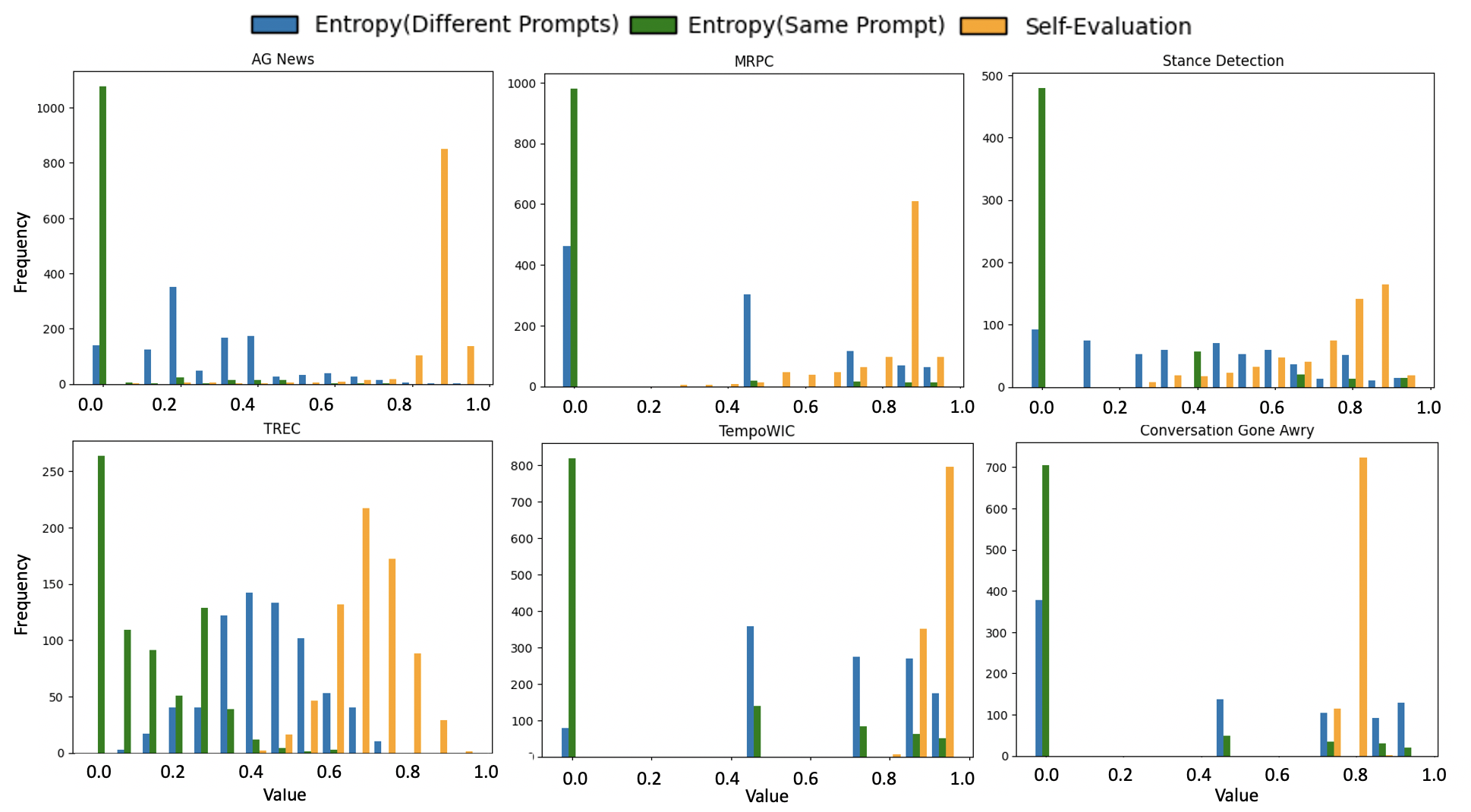}
  \caption{\small Distribution of entropy and confidence values.}%\ella{I will improve its resolution}}
  \label{fig:histogram}
\end{figure*}
\begin{figure*}[h]
  \centering
  \includegraphics[width=\textwidth]{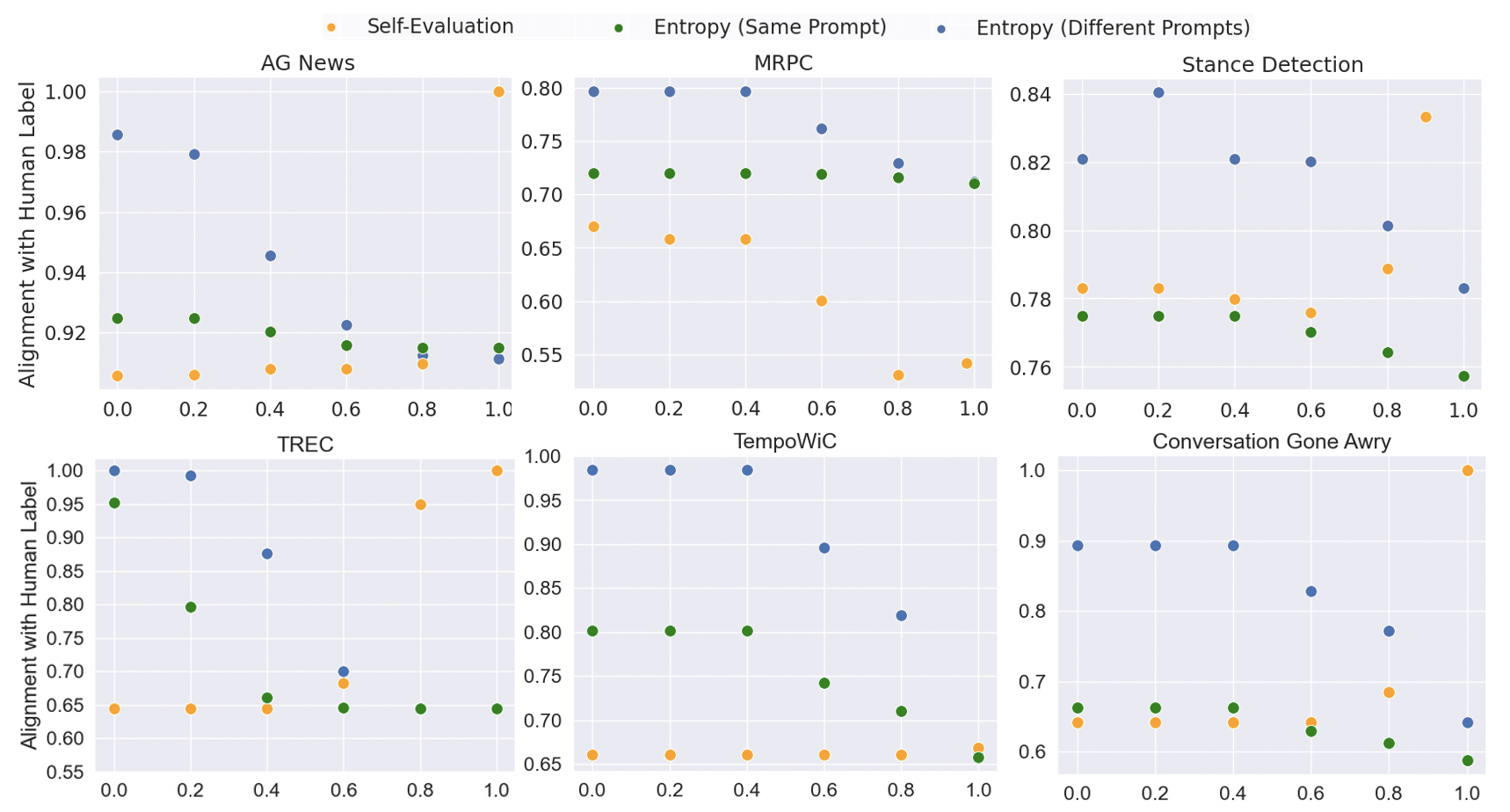}
  \caption{\small Scatter plots of the average alignment of ChatGPT's annotation with human annotation for train data against the threshold. 
  We vary the threshold for different metrics during work allocation to investigate the effectiveness of different metrics in quantifying ChatGPT's annotation capability.}
  \label{fig:threshold}
\end{figure*}

\subsection{LLM Annotation}

We obtain responses from ChatGPT (\texttt{gpt-3.5-turbo}) due to its high-quality annotations and low inference cost \citep{kuzman2023chatgpt} using different prompts carefully crafted in Table \ref{tab:prompt}.
If the response is an ambiguous answer such as \textit{"I cannot determine the class of the text"}, we encode it as a new class label which can result in higher uncertainty metrics. The uncertainty computation decides whether annotation will be finally allocated to ChatGPT, and if so, we decide the final label with a majority vote across ChatGPT's generations \citep{wang2022self}. 

\subsection{Evaluation}
To evaluate the quality of datasets annotated with different strategies, we fine-tune the same RoBERTa base classifier and calculate macro F1 scores on test data for a fair comparison. We report macro F1 as a more accurate representation of the performance due to the unbalanced nature of \llm{}' annotations for some datasets.

In terms of cost, we only consider monetary cost in this work. We calculate human annotation costs based on what was reported in the dataset paper. If the information is not applicable, we assume each instance is annotated by 5 independent annotators with a wage of $\$15/$hour.
We calculate ChatGPT annotation cost using the product of the token length of the input prompt and the price of calling API for (\texttt{gpt-3.5-turbo}) ($\$0.002/1k$ tokens) at the time of experimentation.

\section{Results}
\begin{table*}[h!]
\centering
\def\arraystretch{1.5}
\resizebox{\textwidth}{!}{  
\begin{tabular}{@{}LCCCCCCCLLLLCCLLLLC@{}}
\toprule
                                 & \multicolumn{6}{c}{\textbf{AGNews}}                                                              & \multicolumn{6}{c}{\textbf{TREC}}                                                                                                    & \multicolumn{6}{c}{\textbf{Stance Detection}}                                                                        \\ \midrule
\textbf{\% ChatGPT} & 0    & 20            & 40            & 60            & 80            & \multicolumn{1}{c|}{100}  & 0    & \multicolumn{1}{c}{20} & \multicolumn{1}{c}{40} & \multicolumn{1}{c}{60} & \multicolumn{1}{c}{80} & \multicolumn{1}{c|}{100}  & 0    & \multicolumn{1}{c}{20} & \multicolumn{1}{c}{40} & \multicolumn{1}{c}{60} & \multicolumn{1}{c}{80} & 100  \\
\textbf{Strategies}              & \multicolumn{18}{c}{\textbf{Macro F1}}                                                                                                                                                                                                                                                                                                                   \\
Random                           & 88.2 & 87.9          & 85.8          & 79.8          & 81.8 & \multicolumn{1}{l|}{\textbf{82.6}} & 92.1 & 88.1                      & 86.1                      & 81.6                      & 76.4                      & \multicolumn{1}{c|}{\textbf{75.8}} & 60.2 & 53.9                      & 53.6                      & 55.0                      & 50.4                      & \textbf{53.6} \\
Self-Evaluation                  & 88.2 & 86.0          & 84.9          & 84.1^*          & 82.1          & \multicolumn{1}{l|}{82.1} & 92.1 & 91.5                      & 87.2                      & \textbf{86.5}^*                      & 76.4                      & \multicolumn{1}{c|}{74.3} & 60.2 & 56.9                      & 54.8                      & 54.4                      & 52.8                      & 52.9 \\
Entropy (Diff. Prompts)                          & 88.2 & \textbf{88.4} & \textbf{88.2} & \textbf{87.4}^* & 84.0          & \multicolumn{1}{l|}{\textbf{82.6}} & 92.1 & \textbf{91.9}                     & \textbf{87.4}                      & 80.8                      & \textbf{79.2}                      & \multicolumn{1}{c|}{\textbf{75.8}} & 60.2 & \textbf{58.2}                      & \textbf{55.1}                      & \textbf{56.8}                      & \textbf{54.7}                      & \textbf{53.6} \\ 
Entropy (Same Prompt)                         & 88.2 & 85.1 & 85.5 & 85.4^* & \textbf{84.7}          & \multicolumn{1}{l|}{81.4} & 92.1 & 90.8                      & 87.1                      & 83.7                      & 76.2                      & \multicolumn{1}{c|}{74.0} & 60.2 & 54.2                      & 53.3                      & 54.0                      & 52.1                      & 47.8 \\ \midrule
                                 & \multicolumn{6}{c}{\textbf{TempoWIC}}                                                            & \multicolumn{6}{c}{\textbf{MRPC}}                                                                                                    & \multicolumn{6}{c}{\textbf{Conversation}}                                                                       \\ \midrule
\textbf{\% ChatGPT}          & 0    & 20            & 40            & 60            & 80            & \multicolumn{1}{c|}{100}  & 0    & \multicolumn{1}{c}{20} & \multicolumn{1}{c}{40} & \multicolumn{1}{c}{60} & \multicolumn{1}{c}{80} & \multicolumn{1}{c|}{100}  & 0    & \multicolumn{1}{c}{20} & \multicolumn{1}{c}{40} & \multicolumn{1}{c}{60} & \multicolumn{1}{c}{80} & 100  \\
\textbf{Strategies}                       & \multicolumn{18}{c}{\textbf{Macro F1}}                                                                                                                                                                                                                                                                                                                             \\
Random                           & 57.5    & 55.9             & 53.2             & 46.2             & 50.3        & \multicolumn{1}{c|}{42.0}    & 83.4    & 78.6                      & 74.4                      & 70.3                      & 65.8                      & \multicolumn{1}{c|}{\textbf{65.9}}    & 71.3    & 63.1                      & 54.5                      & 57.1                  & 50.0                      & 54.1   \\
Self-Evaluation                  & 57.5    & 57.8                      & 55.9^*                      & 51.8^*                      & 52.9^*                     & \multicolumn{1}{c|}{\textbf{43.0}}    & 83.4    & 79.7                      & 77.8                      & 71.5                      & 63.9                      & \multicolumn{1}{c|}{58.6}    & 71.3    & \textbf{70.1}^*                      & 62.6                     & \textbf{64.2}^*                      & 50.4                      & 50.7    \\
Entropy (Diff. Prompts)                         & 57.5    & \textbf{58.4}^*             & \textbf{56.9}^*             & \textbf{55.9}^*             & \textbf{53.8}^*        & \multicolumn{1}{c|}{42.0}    & 83.4    & \textbf{80.0}                      & \textbf{79.8}^*                      & \textbf{76.6}^*                      & \textbf{73.1}^*                      & \multicolumn{1}{c|}{\textbf{65.9}}    & 71.3    & 66.5                      & \textbf{64.2}^*                      & 62.6                      & \textbf{56.2}^*                      & 54.1    \\ 
Entropy (Same Prompt)            & 57.5   & 56.3             & 53.5             & 52.9^*             & 43.8        & \multicolumn{1}{c|}{42.0}    & 83.4    & 79.2                      & 72.7                      & 67.7                      &          68.6          & \multicolumn{1}{c|}{65.7}    & 71.3    & 55.4                      & 55.4                      & 54.1                     & 54.8                     & \textbf{54.6}    \\ \bottomrule
\end{tabular}}
\caption{Test performance of fine-tuned RoBERTa under different allocation strategies. We vary the percentage of data allocated to ChatGPT for annotation and carry out finetuning using train data annotated under different strategies for all six datasets. Figure with superscript * means the result under that strategy is significantly better than baseline strategy at 10\% significance level.} 
\label{tab:strategy_performance_comparison}
\end{table*}
\subsection{Strategy Comparison}
We plot the histograms for distribution of uncertainty metrics (entropy with different prompts and same prompt as well as confidence score). From Figure \ref{fig:histogram}, we can observe that the model tends to be confident with its predictions with a skewed distribution towards high confidence value although we ask ChatGPT to normalize its answer.

We hypothesize that a lower level of uncertainty in ChatGPT's response indicates a higher degree of reliability in the label. 
Therefore, we set different thresholds for entropy (lower than an entropy threshold) and self-confidence score (higher than a confidence threshold) to select data that ChatGPT is more certain about. For those instances selected, we evaluate ChatGPT's annotation quality by calculating its alignment with the gold label (human annotation). Figure \ref{fig:threshold}'s decreasing trends for entropy-guided allocation (green and blue dots) on all datasets validate our hypothesis of an inverse relationship between uncertainty and annotation quality. It justifies the helpfulness of using the entropy of ChatGPT's annotations as an estimate for its annotating expertise. Importantly, we observe that ChatGPT's self-reported confidence scores (orange dots) are not consistently a good estimate for its annotation quality. For some datasets such as AG News (top left), most of the data (94.3\% with calculation) has high self-reported confidence ranging from 0.8 to 1, which leads to a weak separation of data in terms of annotation quality. For MRPC (top middle), there is a decreasing trend where data instances with higher confidence scores in fact have a poorer alignment with gold labels. This shows that the reliability of using self-reported confidence from \llm{} is not guaranteed.

\begin{figure*}[h!]
  \centering
  \includegraphics[width=\textwidth]{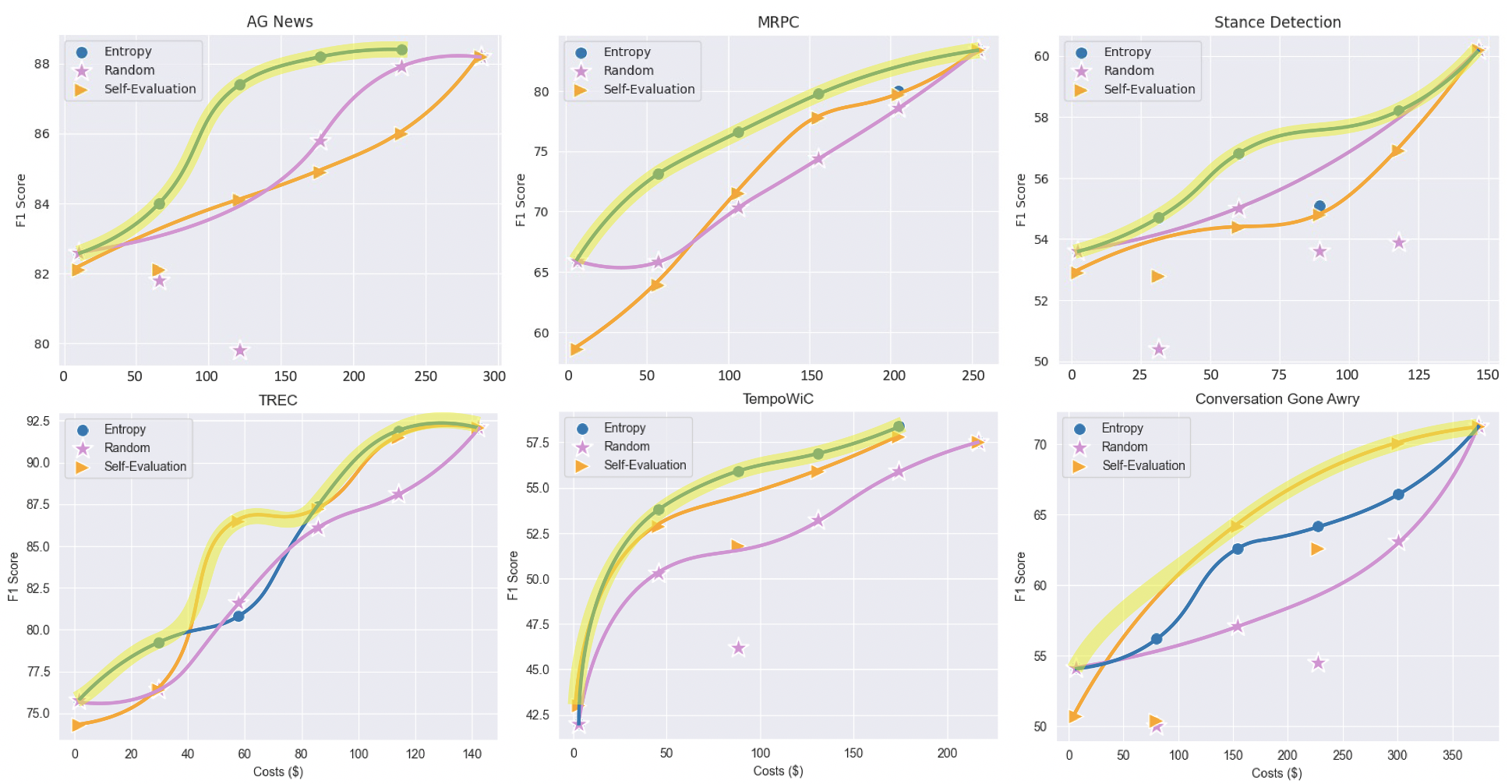}
  \caption{Pareto curves under different allocation strategies (\textcolor{randompink}{random}, \textcolor{entropyblue}{entropy guided}, \textcolor{orange}{self-evaluation guided}). The Pareto frontier is \hl{highlighted}, illustrating the optimal choices that are Pareto efficient.
  }
  \label{fig:pareto}
\end{figure*}

The purpose of achieving a higher quality for train data is to ensure that it can teach the classifier accurate information through fine-tuning. In Table~\ref{tab:strategy_performance_comparison}, we carry out comparisons of different allocation strategies in terms of test performance after fine-tuning with such data labeled. 
We see that holding the proportion of data allocated to ChatGPT fixed (e.g., taking the setup of 40\% for TempoWIC as an example), our proposed uncertainty-guided allocation using self-evaluation and entropy results in a better-annotated dataset,  reflected by its higher test F1 (56.9) than the random allocation baseline (53.2). More often than not, entropy-guided allocation is better than confidence-guided allocation. This is probably due to the skewed distribution of self-reported confidence, resulting in a poorer distinguishability between instances \llm{} are better or worse at annotating. 

\begin{table*}[h!]
\centering
\def\arraystretch{2.25}
\resizebox{1\textwidth}{!}{%
\begin{tabular}{@{}lMll@{}}
\toprule
\multicolumn{1}{l}{\textbf{Dataset}} & \multicolumn{1}{l}{\textbf{Text}} & \multicolumn{1}{l}{\textbf{Groundtruth}} & \multicolumn{1}{c}{\textbf{ChatGPT}} \\ \midrule
\multirow{1}{*}{AG News}         & Title: Sprint Set to Debut Video-Streaming Cell Phone Description: OVERLAND PARK, Kan. (AP) -- Channel surfing is moving off the couch as Sprint Corp...                   &                  Sci/Tech           &                Business               \\
\multirow{1}{*}{TREC}                & What does A\&W of root beer fame stand for?                   &    Abbreviation                                  &               Entity                               \\
\multirow{1}{*}{Stance Detection}         & @user As a former fetus I oppose \#ProlifeYouth \#SemST                   &                        Negative              &              Neutral                         \\
\multirow{1}{*}{Conversation}        & \rule{0pt}{2ex} \textbf{rjoccolenty}: Shouldn't her name be Zainab Yusef and not Zainab Khan?
\textbf{Bluebolt94}: Does the credits at the end of the episode say ''Zainab Yusef''? No they say ''Zainab Khan'' and Yusef called her ''Mrs. Khan'' during the episode. So no, her name is ''Zainab Khan''.  –
\textbf{AnemoneProjectors}: The Khans are clearly not as traditional as the Masoods, or Afia would have been called Afia Yusef. We already know this! And what GS said. Watch the show properly P ––   \rule[-2ex]{0pt}{0pt}                 &      True                                &   False                                           \\
\multirow{1}{*}{MRPC}                & \textbf{Sentence1:} At 5 p.m. EDT , Henri had maximum sustained winds near 50 mph , with some gusts reaching 60 mph. \textbf{Sentence2:} At 8 p.m. Friday , Henri was becoming disorganized , but still had maximum sustained winds near 50 mph , with stronger gusts.   \rule[-1.5ex]{0pt}{0pt}               &              Not paraphrase                   &         Paraphrase                                     \\
\multirow{1}{*}{TempoWiC}            & \textbf{tweet 1}: If you need some to watch on Netflix, containment is so good. \textbf{tweet 2}: I have a lot of questions about the containment series. \textbf{target word}: containment                   &      Same                                &    Different                                         \\ \bottomrule
\end{tabular}}
\caption{Specific instances with high entropy values for ChatGPT annotations.}
\label{tab:qual}
\end{table*}

\subsection{Pareto Efficient Allocation}
%\diyi{figure 4 is not referred.}
By plotting test performance against annotation cost (Figure~\ref{fig:pareto}), practitioners can visualize the tradeoff in annotation quality achievable at different budgets with collaboration between human and an LLM like ChatGPT by studying the Pareto frontier. Points along the highlighted Pareto frontier mean it is theoretically impossible to achieve a higher test accuracy without increasing the budget, and it is also impossible to reduce the cost but achieve the same level of annotation quality. Furthermore, it provides information on the approximate proportion that can be outsourced to ChatGPT to achieve human-level performance. For more straightforward tasks like topic classification, part of the annotation work could be potentially outsourced to ChatGPT and lead to a cost reduction (e.g., AG News: 33\%) by ensuring human-level annotation performance. For datasets requiring nuanced comprehensions like Stance Detection and Conversation Gone Awry, any level of 
outsourcing to the current version of ChatGPT compromises annotation quality. 
Practitioners can choose among the Pareto efficient points based on their budgets.

\subsection{Qualitative Analysis}
We select some instances with entropy values higher than 0.8 from each dataset (Table~\ref{tab:qual}) to understand the current challenges faced by ChatGPT in annotating data. We find that ChatGPT has high uncertainty for instances containing sarcasm and incomplete sentences that require more inference during opinion mining. For example, in deciding the stance towards abortion for the tweet ``\textit{as a former fetus I oppose}'', the incomplete nature of this sentence causes confusion to ChatGPT. Also, it struggles with numerical reasoning as seen from its inability to compare wind speed during paraphrase detection and may be misled by some keywords (``Corp'') related to other incorrect classes (``business'') in topic classification.
\subsection{Ablation Study}
\label{ablation}
We carry out inferences with the same  instruction-formatted prompt for the same number of times and compute the entropy for ChatGPT's responses. From Figure~\ref{fig:threshold}, we observe some extent of effectiveness of computing entropy using the same prompt in quantifying ChatGPT's capability, as reflected by a decreasing pattern of alignment with the increased threshold. However, it serves as a much weaker method to quantify expertise compared with our method with different prompt designs since the majority of the data has zero entropy (see Figure~\ref{fig:histogram}). This suggests that ChatGPT's responses are generally consistent within multiple applications of the same prompt. In Table~\ref{tab:strategy_performance_comparison}, the test performance of entropy-guided allocation under different prompts is consistently higher than when based on a single prompt. The performance gap gives strong evidence of the utility of applying different prompt types in Table \ref{tab:prompt}.

\section{Conclusion}
This work introduces \textit{CoAnnotating}, a framework which takes a collaborative angle to view the relationship between humans and \llm{} when annotating each dataset. Under this framework, we use uncertainty metrics to estimate \llm{}' annotating capability and guide effective work allocation. Moreover, we apply the Pareto efficiency concept for practitioners to compare strategies and understand cost-performance tradeoffs. The empirical results demonstrate the effectiveness of our proposed framework in achieving greater cost efficiency. Overall, our framework provides important insights around the reliability of self-reported confidence score by \llm{}, the sensitivity of ChatGPT's responses to prompt variations as well as the extent to which human resources can be freed by \llm{} to be put on more meaningful areas.

\nocite{Ando2005,borschinger-johnson-2011-particle,andrew2007scalable,rasooli-tetrault-2015,goodman-etal-2016-noise,harper-2014-learning}

\section{Limitations}
Since \llm{} has been trained on a large number of datasets, there may be data leakage issue where \llm{} has seen some datasets in our experiment, making entropy values obtained for \llm{}' responses lower. As an initial exploration of the co-annotating concept, this work aims for human-level performance in annotating datasets. It does not consider the scope of superhuman-level performance where we treat human annotation in each dataset as gold labels. Future work can further investigate the instances where \llm{} actually annotates better than humans. We consider annotating profiles of human and \llm{} as two groups but this framework can be further enriched by taking variations within each group (expert, crowd workers, different \llm{}) into considerations. More exploration can also be carried out to investigate how to design prompts in a way that can increase \llm{}'s annotating expertise so that more annotation work can be outsourced to \llm{} for greater cost efficiency. Moreover, this work only did experiments for classification tasks and English datasets. However, the idea of \textit{CoAnnotating} is generalizable to generation tasks and datasets in other languages as well, which are meaningful to study in future work.

\section*{Ethical Statement}
We are aware of the potential ethical concerns of using \llm{} as potential labelers in the data annotation process in terms of the perpetuation of existing biases in \llm{}. Since \llm{} are trained on vast amounts of texts on the Internet, they can unavoidably incorporate the biases present in these data sources. Such biases could be under-representation of certain demographic groups, cultural stereotypes as well as linguistic biases. However, we believe that the benefit of proposing a collaborative co-annotation framework outweighs the potential risks related to the framework.

\section*{Acknowledgement}
We are thankful to the members of SALT Lab and WING Lab as well as anonymous EMNLP reviewers for
their helpful feedback. Minzhi Li is
supported by the A*STAR Computing and Information Science (ACIS) Scholarship. Caleb Ziems is supported by the NSF Graduate Research Fellowship under Grant No. DGE-2039655. We would like to acknowledge a grant from the Office of Naval Research to DY,  and National Research Foundation, Singapore under its AI Singapore Programme (AISG Award No: AISG2-GC-2022-005).

\bibliography{main}
\bibliographystyle{acl_natbib}

\include{appendix}

\end{document}